# Overview of GeoLifeCLEF 2023: Species Composition Prediction with High Spatial Resolution at Continental Scale Using Remote Sensing

Christophe Botella[1], Benjamin Deneu[1,2], Diego Marcos[1], Maximilien Servajean[3], Théo Larcher[1], César Leblanc[1,2], Joaquim Estopinan[1,2], Pierre Bonnet[4] and Alexis Joly[1]

[1]*INRIA, LIRMM, Univ Montpellier, CNRS, Montpellier, France*
[2]*AMAP, Univ Montpellier, CIRAD, CNRS, INRAE, IRD, Montpellier, France*
[3]*LIRMM, AMIS, Univ Paul Valéry Montpellier, Univ Montpellier, CNRS, France*
[4]*CIRAD, UMR AMAP, Montpellier, Occitanie, France*

**Abstract**
Understanding the spatio-temporal distribution of species is a cornerstone of ecology and conservation. By pairing species observations with geographic and environmental predictors, researchers can model the relationship between an environment and the species which may be found there. To advance the state- of-the-art in this area with deep learning models and remote sensing data, we organized an open machine learning challenge called GeoLifeCLEF 2023. The training dataset comprised 5 million plant species observations (single positive label per sample) distributed across Europe and covering most of its flora, high-resolution rasters: remote sensing imagery, land cover, elevation, in addition to coarse-resolution data: climate, soil and human footprint variables. In this multi-label classification task, we evaluated models ability to predict the species composition in 22 thousand small plots based on standardized surveys. This paper presents an overview of the competition, synthesizes the approaches used by the participating teams, and analyzes the main results. In particular, we highlight the biases faced by the methods fitted to single positive labels when it comes to the multi-label evaluation, and the new and effective learning strategy combining single and multi-label data in training.

## 1. Introduction

Land use changes and global warming transform ecosystems at an alarming rate ([1]), but their local impact on biodiversity is highly context-dependent and difficult to predict. Regularly monitoring species composition at high spatial resolution (≈50 m) and continental or global extent would enable to understand in real time how species communities and biodiversity indicators (e.g. diversity, habitat condition, presence of endangered species) respond to global changes in order to take effective actions, but it is largely unfeasible. Nevertheless, species

---

christophe.botella@inria.fr (C. Botella)

composition predictions across space and time from deep learning-based species distribution models (deepSDMs, [2, 3]) are a promising alternative as these models can efficiently exploit complex and high spatial resolution geographic predictors, including remote sensing data, to fill the sampling gaps ([4, 5]). However, the scarcity, imbalance and heterogeneity of the available species observations and environmental data are major impediments to the implementation of species distribution models at this resolution.

Standardized biodiversity observation data, such as presence-absence surveys in small plots, can only cover a very limited spatial extent and is costly to renew. New biodiversity monitoring schemes, such as crowdsourcing programs (e.g. Pl@ntNet, iNaturalist, Observation.org) are a complementary data source, as they provide millions of presence-only (PO) species records with precise geo-location every year. However, PO records do not inform about the local absence of the non-observed species ([6]), only show a small portion of species communities in under-sampled areas, are biased towards certain species, and thus carry many biases when used in species distribution models (see e.g. [7, 8]). Therefore, the evaluation of species distribution models on PO data induces important evaluation biases only due to the sampling patterns, as pointed out by the previous GeoLifeCLEF campaign ([9]). This highlights the need for an evaluation procedure based on exhaustive sampling of species communities at high spatial resolution. In addition, a small number of standardized species observations, e.g. presence-absence (PA) plots, can also help solve many sampling biases of the PO data when jointly integrated into species distribution model calibration, while enabling to capitalize on the rich information hidden in the mass of PO data ([10, 11]). Even when using comprehensive PA data, it is difficult to model and map biological groups with large taxonomic diversity such as plants, i.e. more than 10 thousand species in Europe, with a few species being very common and a vast majority of them being rare. This problem is referred to as strong class imbalance in the domain of machine learning.

Remote sensing data is a central resource to characterize environment at high spatial or temporal resolution, but its integration into species distribution models is quite recent ([4]). This environmental data is precious to complete the environmental landscape at coarser spatial scale, as characterized by climatic, soil or land cover descriptors, but the different spatial resolutions makes it difficult to integrate in classical deep learning architectures.

For this new open model evaluation campaign, called GeoLifeCLEF 2023, we have assembled an open dataset at a European scale to investigate these problems and designed the evaluation of the multi-label prediction of species composition at high spatial resolution based on standardized PA data for the first time in GeoLifeCLEF. To summarise, the difficulties of the challenge include: multi-label learning from a large amount of single positive labels and a small number of partial multi-label samples, strong class imbalance, large-scale, learning from multiple type of predictors, including multi-band satellite images and time-series.

## 2. Dataset and Evaluation Protocol

We briefly describe below the dataset of GeoLifeCLEF 2023, i.e. the species observations and environmental predictors made available to train models, and its evaluation protocol, namely the standardized PA data used for evaluation and evaluation metric. We acknowledge that a more detailed description of the dataset components, structure (e.g. files and links between them) and generation protocol is provided in the dedicated data paper ([12]). The competition was hosted by Kaggle[1].

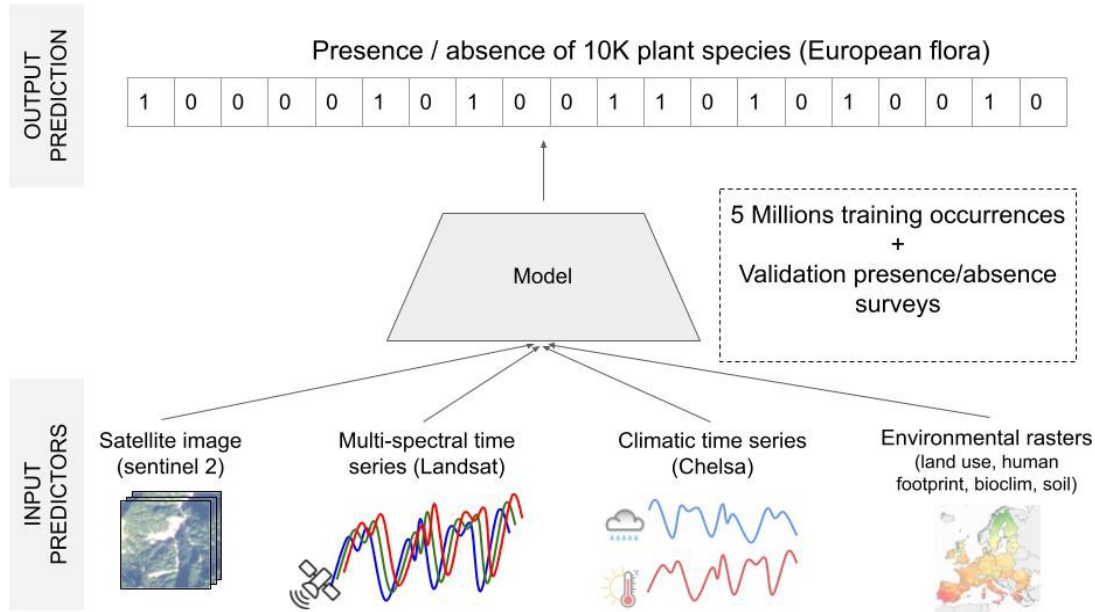

**Figure 1:** GeoLifeCLEF 2023 aimed at developing and evaluating models that predict plant species composition at high spatial resolution (~10m) from diverse type of input environmental predictors, by calibrating them on two types of species observations: Opportunistic presence-only records and standardized presence-absence surveys.

**Observation data.** The training species observation data was composed, on the one hand, of more than 5 million plant species PO records (species name, geo-location, time, etc.) and, on the other hand, of around 5.9 thousand **presence-absence** (PA) surveys, all collected between 2017 and 2021, and with a geo-location uncertainty under 100 m. The PO data was extracted from the Global Biodiversity Information Facility (GBIF, [13]), combining 13 trusted source datasets, including three international citizen science programs (Pl@ntNet, iNaturalist RG, Observation.org) and regional datasets, favouring a large spatial coverage across Europe (38 countries). Each PA survey consisted in the exhaustive inventory of all plant species in a small plot of 10 to 400m². The PA data was composed of four source datasets covering France and Great Britain, namely the "Données de l'inventaire forestier national de l'IGN", the "National Plant Monitoring Scheme (Great Britain)", the "Conservatoire Botanique National Méditerranéen" and the "Conservatoire Botanique National Alpin". The PO and PA data overall covered 10,038 species, i.e. most of the European flora. This species observation data was provided to inform the model outputs, keeping in mind that the PO data only partially represented the local species composition and was subject to various sampling biases. Indeed, we recall that a PO record of one species does not inform about the absence of other species: an observer might not have reported another species because it was difficult to detect at this time, to identify, or because it was simply not interesting to this observer. The PA surveys were provided to control for the sampling biases in model calibration.

---
[1]www.kaggle.com/competitions/geolifeclef-2023-lifeclef-2023-x-fgvc10

**Environmental predictors.** We also provided spatialized geographic and environmental data to be used as predictors, i.e. input for model predictions. For each species observation (in PO and PA), we provided a four-band 128x128 **satellite image** at 10 m resolution around the occurrence location and **satellite quarterly time series** of the past values for six spectral bands at the location (over 20 years). Besides, we provided various **environmental rasters** at the European scale, including climatic, soil, land cover, human footprint and elevation variables. These three types of input environmental predictors are illustrated in Figure 2. We also provided monthly rasters of four climatic variables from which it was possible to extract time series of climatic variables for any observation. All predictors are summarized in Table 1 along with their source and spatial resolution.

**Train-test split.** A spatial block hold-out on a grid of width and height of 50 km was performed to split the PA surveys into a validation set of 5,948 surveys ($\approx$ 20%), provided for model training, and a test set of 22,404 surveys ($\approx$ 80%), for which the sampled species where hidden from the participants. The location of the validation and test surveys is represented in the map of Figure 3. This spatial block hold-out split strategy is employed to limit the effect of the spatial auto-correlation in the data when evaluating the models and assess their ability to extrapolate to new regions [14].

**Evaluation metric.** GeoLifeCLEF 2023 was proposed as a multi-label classification task. The main evaluation metric for the Kaggle competition was the micro $F_1$-score computed on the PA test set. The $F_1$-score is a measure of overlap between the predicted and actual set of species present, averaged over the test PA surveys. Each survey $i$ is associated with a set of ground-truth labels $Y_i$, i.e. the set of plant species present at $i$. For each survey, participants provided a list of predicted labels $\hat{Y}_{i,1}, \hat{Y}_{i,2}, \ldots, \hat{Y}_{i,\hat{R}}$. The micro $F_1$-score is then computed using

$$F_1 = \frac{1}{N} \sum_{i=1}^{N} \frac{\text{TP}_i}{\text{TP}_i + (\text{FP}_i + \text{FN}_i)/2}$$

Where 
$$\begin{cases} \text{TP}_i = \text{Number of predicted labels truly present, i.e. } |\hat{Y}_i \cap Y_i| \\ \text{FP}_i = \text{Number of labels predicted but absent, i.e. } |\hat{Y}_i \setminus Y_i| \\ \text{FN}_i = \text{Number of labels not predicted but present, i.e. } |Y_i \setminus \hat{Y}_i| \end{cases}$$

The micro $F_1$-scores are reported for the 19 main methods tested in the challenge in Table 2. We also provided in this table the "Score sp.", a macro-averaged $F_1$ score where true positive, false positive and false negative are counted per species to compute its score, which is then averaged over species.

## A. Satellite image patches

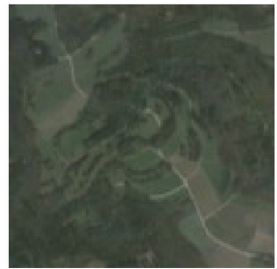
Red Green Blue (RGB)

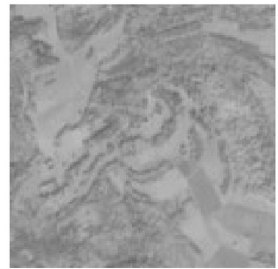
Near Infra-Red (NIR)

## B. Satellite time series

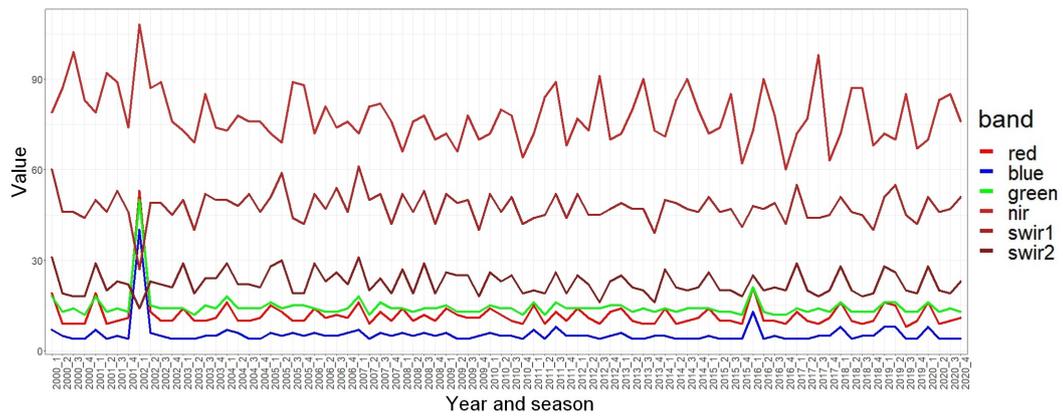

## C. Environmental rasters

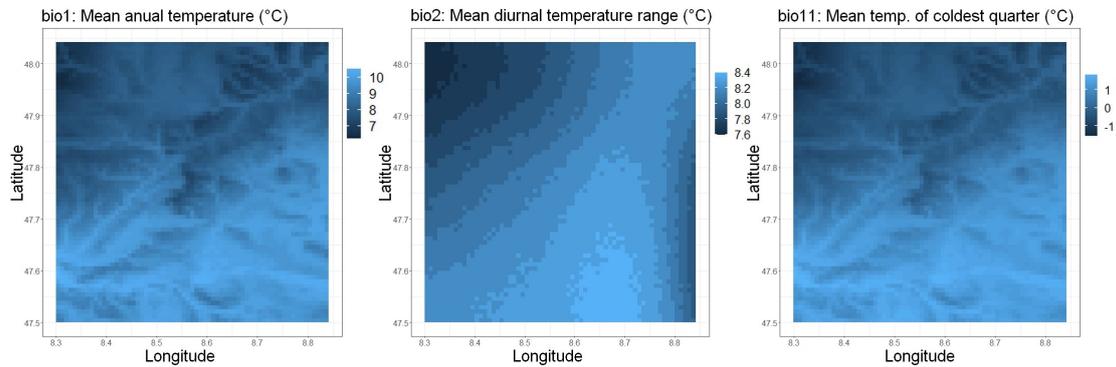

**Figure 2:** Illustration of the environmental data for an occurrence collected in northern Switzerland (lon=8.5744;lat=47.7704) in 2021. A. The 1280x1280m satellite image patches sampled in 2021 around the observation. B. Quarterly time series of six satellite bands at the point location since winter 1999-2000.
C. Three example bioclimatic images (65x65km) around the observation, extracted from the provided environmental rasters.

| Name | Description | Source | Resolution |
|---|---|---|---|
| Climate | 19 rasters of historical bioclimatic variables (1981-2010) traditionally used in SDMs | CHELSA | ~1 km |
| Montly Climate | 4 variables from January 2000 to december 2019 | CHELSA | ~1 km |
| Soil | 9 pedological rasters | Soilgrids | ~ 1 km |
| Elevation | Elevation above sea level | ASTER | ~ 30 m |
| Land cover | According to IGBP classification (17 classes) | MODIS 500 m | ~ 500 m |
| Human footprint | 7 pressures on the environment for 1993 and 2009 | Venter et al., 2016 | ~ 1km |
| Satellite imagery | RGB and NIR patches centered on each observation and taken the same year | Sentinel-2 | 10 m |
| Satellite time series | Time series of six quarterly satellite bands values since winter 1999 | Landsat | 30 m |

**Table 1**
Summary of the environmental predictors associated with the species observations, source and spatial resolution.

## 3. Participants and methods

Seven participants from four countries participated in the GeoLifeCLEF 2023 challenge and submitted a total of 130 entries (i.e. runs, see 4). In Table 2, we report the performance achieved by the main documented methods of the participants as well as the baseline methods that we developed. Hereafter, we briefly describe those different methods:

### 3.0.1. Participants' methods

- **KDDI research**: This team trained various convolutional neural networks (CNN), all based on the ResNet backbone (ResNet34 and 50). One of the CNNs was trained solely on the 19 bioclimatic rasters, while others were multi-modal networks with a late fusion layer to merge the different modalities used (see Table 2). The best performing run was an ensemble of the best models based on a simple average of their output. The best models were trained in three steps, firstly on the PA plots with a binary cross-entropy loss, then fine-tuned on the PO records with a cross-entropy loss, and finally fine-tuned again on the PA with the binary-cross-entropy loss. This team carried an ablation study showing the importance of these three steps. The detailed methodology of this team is explained in their working note ([15]).

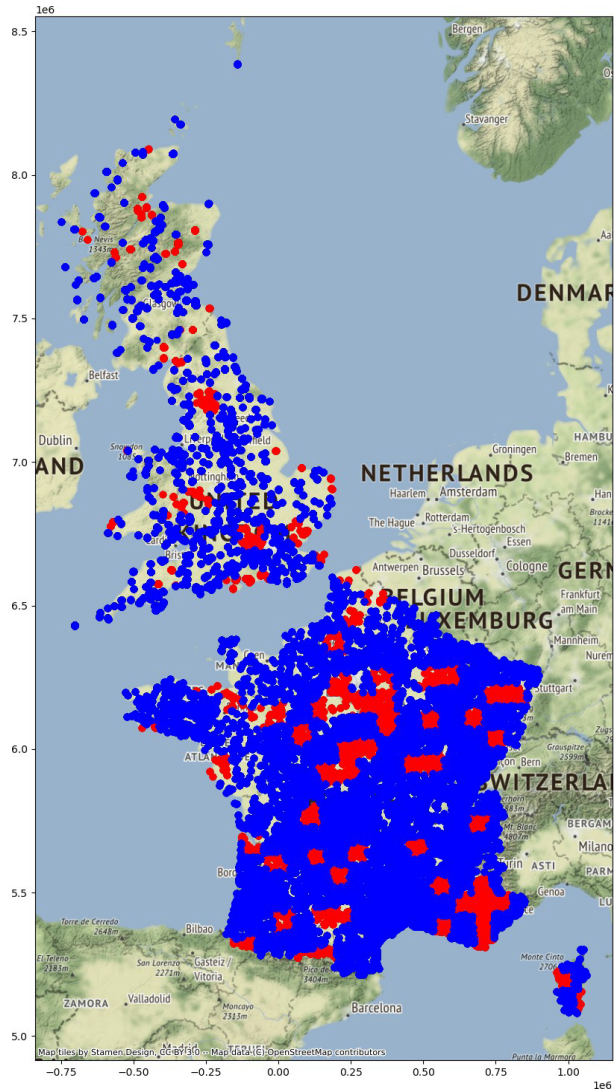

**Figure 3:** The locations of the validation (red) and test (blue) presence-absence (PA) surveys. The 22 thousand test PA labels were hidden from the participants and used for model evaluations, while the five thousand validation PA were provided to them to help control sampling biases in model calibration.

- **Jiexun Xu**: This researcher focused on the tabular environmental data only, i.e he did not use the spatial structure of the environmental co-variates nor the remotes sensing images and times series. The model used is XGBoost, which was trained on the PA plots only based on the following predictors: climate, soil, land cover and the detailed human footprint variables. He also added the one-hot encoded species presences in GBIF in a 1km radius of these plots as input variables.
- **Lucas Morin**: This researcher optimized a K-Nearest Neighbor predictor using only the spatial coordinates and the PA plots.
- **QuantMetry**: This team trained various models on the PA data, and their one was a ResNet50 using only the Sentinel2 satellite images (RGB+NIR) as input. The model was pre-trained on the satellite images in a prior work and fine-tuned to the PA data in the challenge.
- **Nina van Tiel**: This researcher used a small CNN, with two convolutional layers and two fully connected layers on the RGB images, along with the bioclimatic, soil and land-cover rasters, trained on the PA plots.
- **Ousmane Youme**: This researcher focused solely on the Landsat time series data at the location of the PA plots. He used a Conv1D neural network model with a binary-cross entropy loss. A common probability threshold was used to convert the predicted species-wise presence probabilities into a set of predicted species.

### 3.0.2. Organizer's baselines

- **MAXENT**: It is a modeling approach widely used in ecology to predict the distribution of a given species based on tabular predictors. The model creates a pre-defined set of non-linear transformations of the environmental predictors consistent with the theoretical ecological response of species to environmental gradients (e.g. quadratic and threshold response functions, see [16]). The statistical model is equivalent to Poisson regression modeling the count of a species per location ([17]). We fitted one MaxEnt model per species present in the PA plots. The species count was set to one when present or zero otherwise. The predictors included were the climate, soil, land cover and the detailed human footprint variables, but only a subset of these variables were included for species with a small number of observations. One random subset of the provided PA plots was used to train all species models. After a fit, the coefficient to transform MaxEnt intensity function into a presence-probability (through the cloglog transform) was calibrated on the training set per species, and a F1 score was computed on the remaining plots per species. We then computed the F1-micro score on this sub-validation set as a function of the number of species kept for prediction by decreasing species-wise F1, the left-out species being always predicted absent. We thus determined that it was optimal to keep only the 391 most trustable species models in prediction (MaxEnt best sp. in Table 2). To aggregate the predicted species-wise presence probability into a predicted set, we selected the $S$ species with highest probability, where $S$ was the rounded sum of probabilities over all kept species. We also submitted the predictions based on all species models.
- **Environmental Random Forest**: Random forests are also widely used in species distribution modeling based on a set of tabular environmental predictors. The Env. Random

Forest models were trained on the same predictors as MaxEnt and on the PA plots. One Random Forest was trained per species found among PA plots and its hyper-parameters were optimized through a cross-validation grid search. The procedure to aggregate the predicted species probabilities into a predicted set was the same as for MaxEnt.

- **Spatial Random Forest**: This version of the Random Forest used only the spatial coordinates of the PA plots as predictors. The procedure to aggregate the predicted species probabilities into a predicted set was the same as above.
- **PA species co-occurring with nearest PO species**: Conditionally to the presence of each species, we computed the proportion of presences of all other species among the PA plots. Then, for each test location, we combined the species probabilities conditionally to the species observed in the PO data in a 1 km radius into a predicted species set through a weighted average. A given species weight was its marginal proportion of presence among validation PA plots.

- **Constant predictor**: This baseline always predicts the same set of species, i.e. the $K$ species that are the most frequent in the validation PA data and where $K$ maximizes the micro-F1 score on the validation PA (top-25 species).
- **Spatial kNN PO**: This baseline implements a k-nearest neighbor method based on the spatial coordinates of the PO data (longitude and latitude). All the species present in the k-nearest neighbors are returned as present. Several values of k (30, 50, 100 and 500) were evaluated on the public leaderboard and only the best one (k=100) is reported here.

## 4. Global results of the evaluation

KDDI research got the best performance of the challenge (best score=0.273 with *KDDI Ensemble*, Table 2), with their ensemble of multi-modal CNNs trained on the combination of PO and PA data. Their best runs had a considerably larger performance than the following group of participants, whose best performances were comparable: Jiexun Xu (2nd, best score=0.226), L. Morin (3rd, best score=0.217), QuantMetry (4th, best score=0.208). These three teams achieved a similar best performance by using only the PA data, but they used very different models, i.e. respectively boosted trees using the environmental variables, a purely spatial nearest neighbor predictor and a CNN based on satellite images. The results of these three teams were also comparable, even though slightly below, to the best baseline method: Maxent best sp. PA (score=0.228, Table 2). The two remaining participants that documented their methods (N. van Tiel, 5th with max. score=0.16; O. Youme, 6th with max. score=0.136) had a lower score than the constant predictor baseline (score 0.162), and their best performance was achieved using CNNs, whose size was smaller in comparison to the ones of other participants who used this such models.

Different participants (i.e. *KDDI team*, *L. Morin* and organizers) reported that, despite the size of the PO data, calibrating models on this data alone did not work well. For instance, our Nearest Neighbors method based on the PO data performed poorly (baseline Spatial KNN PO, score=0.056, Table 2) compared to the Nearest Neighbors method based on the PA data submitted by L. Morin (Morin KNN PA, score=0.217). Besides, the multimodal CNN of *KDDI Research* performed poorly when it was initially calibrated on the PA and then fine-tuned on the PO (KDDI PA/PO, score=0.073) or solely calibrated on the PO (KDDI PO, score=0.058), while it performed well when fine-tuned on the PA (KDDI PO/PA, score=0.223, and KDDI PA/PO/PA, score=0.251).

| Team | Method name | Model | Predictors | Score | Score Sp. |
|---|---|---|---|---|---|
| KDDI | KDDI Ensemble | Ensemble of CNNs | Satellite rgb, nir, soil, climate, human | 0.273 | 0.028 |
| KDDI | KDDI PA/PO/PA | Multi-modal CNN | Satellite rgb, nir, soil, climate, human | 0.251 | 0.031 |
| Baseline | Maxent best sp. PA | Poisson regression | climate, soil, land cover, human | 0.228 | 0.016 |
| JiexunXu | XG-Boost enviro. PA | XG-Boost | climate, soil, land cover, human | 0.226 | 0.030 |
| KDDI | KDDI PO/PA | Multi-modal CNN | Satellite rgb, nir, soil, climate, human | 0.223 | 0.019 |
| L. Morin | Morin KNN PA | Nearest Neighbors | lon/lat | 0.217 | 0.017 |
| Quantmetry | Satellite CNN | ResNet50 | Satellite RGB-NIR | 0.208 | 0.022 |
| KDDI | KDDI PA | Multi-modal CNN | Satellite rgb, nir, soil, climate, human | 0.198 | 0.022 |
| Baseline | Spatial RF PA | Random Forest | lon/lat | 0.193 | 0.006 |
| Baseline | Enviro. RF PA | Random Forest | | 0.191 | 0.005 |
| Baseline | PA assemblage from nearest PO | Nearest Nei + Cooccurre | lon/lat | 0.167 | 0.014 |
| Baseline | Constant PA | Constant predictor | None | 0.162 | 0.002 |
| N. van Tiel | van Tiel's CNN | CNN | Satellite rgb, soil, climate, human | 0.160 | 0.004 |
| O. Youme | time-series CNN PA | CNN | Time-series | 0.136 | 0.021 |
| Baseline | Maxent all sp. PA | Poisson regression | climate, soil, land cover, human | 0.106 | 0.024 |
| KDDI | KDDI PA/PO | Multi-modal CNN | Satellite rgb, nir, soil, climate, human | 0.073 | 0.021 |
| KDDI | KDDI PO | Multi-modal CNN | Satellite rgb, nir, soil, climate, human | 0.058 | 0.014 |
| Baseline | Spatial KNN PO | Nearest Neighbors | lon/lat | 0.056 | 0.032 |

**Table 2**
The 18 main documented results of GeoLifeCLEF 2023 ordered by decreasing main score (F1 micro) - the acronyms PA and PO respectively stand for Presence Absence, i.e. the validation multi-label data was used to fit the model, and Presence-Only which means that the single label data was used.

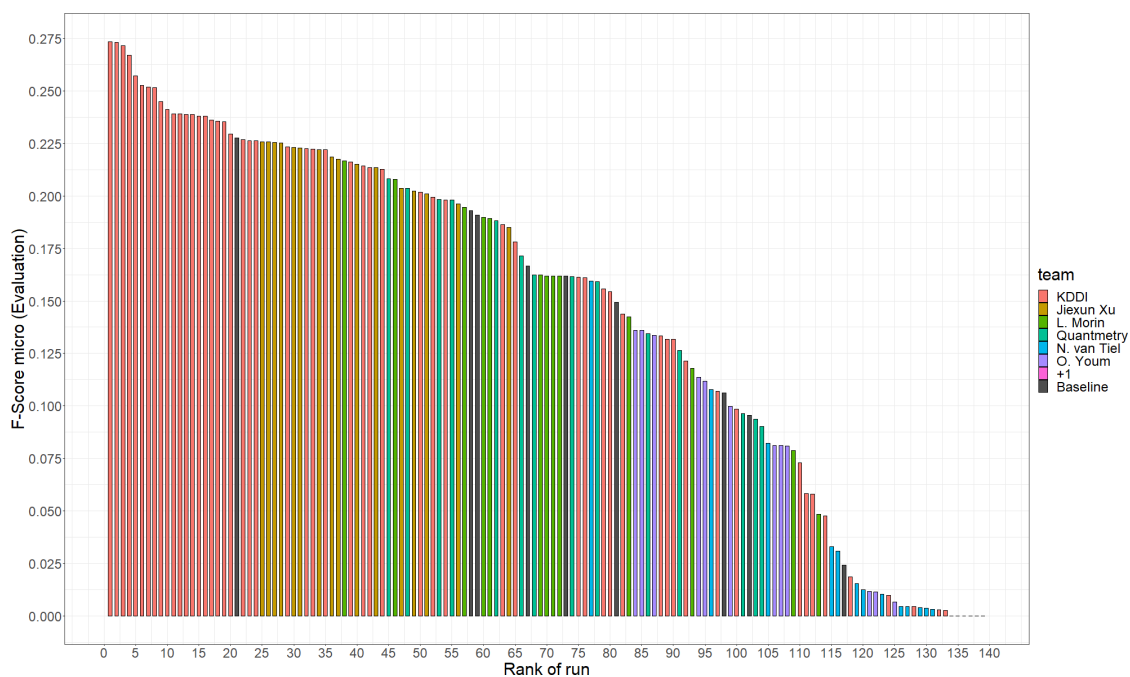

**Figure 4:** Results (F1-micro) ranked by decreasing order for the 139 valid runs submitted to GeoLifeCLEF 2023, including 9 main organizer baselines.

Still, the PA data alone had only a few records for thousands of species. One of the symptoms of this is that the Maxent baseline including all species models (Maxent all sp. PA, score=0.106) performed poorly compared to its version where species models were filtered based on their predictive performance (Maxent best sp. PA, score=0.228). Furthermore, **KDDI research** drastically increased the performance of their multimodal CNN by combining the PA and PO data in its training (KDDI PA/PO/PA, score=0.251) compared to its training on the PA data alone (KDDI PA, score=0.198). Besides, this team showed again that model ensembling can improve performances with their best run (KDDI ensemble, score=0.273, Table 2).

## 5. Complementary analysis

**A strong class detection bias in PO data.** The number of species presences in the PO and PA data were totally unrelated (Figure 5), even when restricting the PO to the same area sampled in the PA data. As a direct consequence, when generating a predictive method to the PO data (e.g. *KDDI PO*), one tends to over-estimate the commonness of the most observed species compared to their commonness in PA surveys. It appears as the main explanation of the performance collapse of *KDDI PO* and *KDDI PA/PO* (scores=0.058 and 0.073, Table 2) compared to *KDDI PO/PA* and *KDDI PA/PO/PA* (scores=0.223 and 0.251).

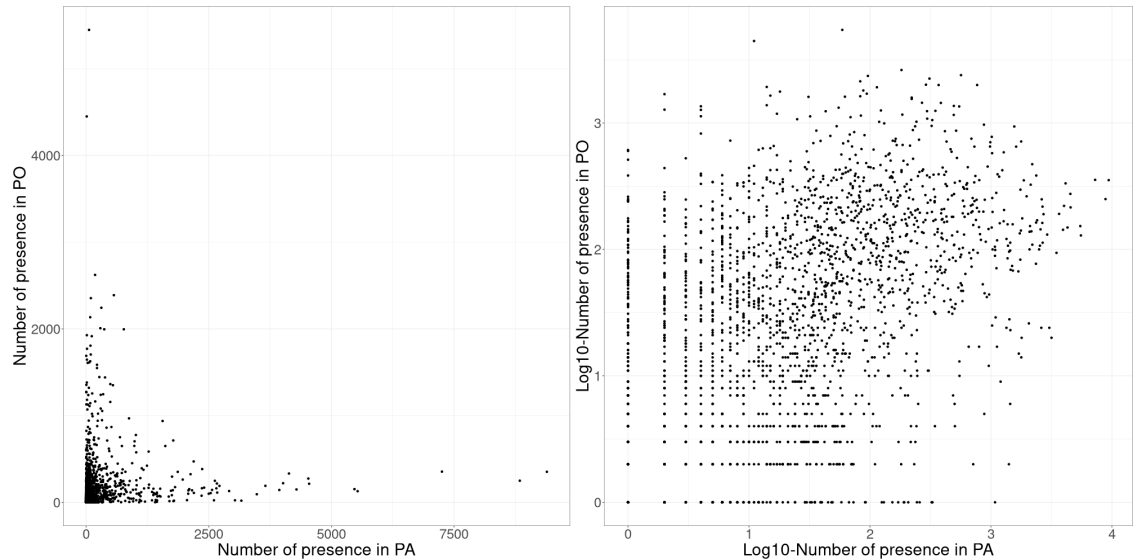

**Figure 5:** Number of presences in the presence-absence (PA) data versus number in the presence-only (PO) data for all the species present in the PA data, in the natural (left) and decimal logarithmic scales (right). Each species is one point. The number of presences in the PA surveys includes validation and test surveys. The PO data are here restricted to a radius of 1km around these PA surveys, so that these numbers reflect the same sampled habitats.

**Room for improvement in set size control.** Another source of error when using the PO data is the predicted set size, because a local set of PO data generally under-represent local species communities due to false absences. It appeared that few participants explicitly optimized set size error in validation. For instance, all methods of the KDDI team predicted a constant set size of 20, which was, on average, 6.5 species more than the actual test species average set size (Bias of their best run *KDDI Ensemble* in Table 3). Hence, *KDDI Ensemble* was less accurate for set size (Abs. error=10.4, Table 3) compared to simpler methods (Abs. error=6.7, 8.2 and 8.3 for resp. *MaxEnt best sp. PA*, *Enviro. RF PA*, *Spatial RF PA*) whose workflow accounted for this aspect. An extreme case is the baseline method **Spatial KNN PO** which used the set of species found among the 100 nearest PO records as prediction, inducing a large positive bias (Bias of 58.25). However, even using The PA data for model training or validation doesn't necessarily induce a control on set size error. For instance, the predicted sets of the baseline method *MaxEnt all sp. PA* were far too large (Bias of 49.9, see Table 3), while restricting the modeled species to the ones with the best to predictive performance resulted in the best error on set size (Bias of -2.3 and Abs. error of 6.7 in Table 3) and a far better main score (*MaxEnt best sp. PA*, score=0.223, see Table 2). The method to aggregate species-wise predicted probabilities into a predicted set might importantly impact set size error. For instance, **the Time-series' CNN PA**, developed by O. Youm, used a binary-cross entropy loss and a single threshold on the species-wise predicted probability of presence to determine the predicted set of species. It resulted in a much larger set size error (Abs. error 15.1 in Table 3) compared to the PA methods that took the sum of the species-wise presence probabilities as the predicted set size such as **MaxEnt best sp. PA** (Abs.

error of 6.7), *Enviro RF PA* (Abs. error of 8.2) and *Spatial RF PA* (Abs. error of 8.3).

| Team | Method name | Abs. error | Bias |
|---|---|---|---|
| Baseline | MaxEnt best sp. PA | 6.688 | -2.273 |
| Baseline | Enviro. RF PA | 8.206 | 2.026 |
| Baseline | Spatial RF PA | 8.284 | 1.394 |
| Quantmetry | Satellite CNN PA | 9.437 | 2.140 |
| Baseline | PA assemblage from nearest PO | 9.771 | 5.079 |
| KDDI | KDDI Ensemble | 10.435 | 6.546 |
| Jiexun Xu | XG-Boost enviro. PA | 12.957 | 8.098 |
| Baseline | Constant PA | 14.449 | 12.546 |
| O. Youm | time-series CNN PA | 15.151 | 10.930 |
| Baseline | Maxent all sp. PA | 50.555 | 49.949 |
| Baseline | Spatial KNN PO | 58.29 | 58.25 |

**Table 3**
Table of set size errors for a subset of the methods whose set prediction procedure was documented. Abs. error is the average (over surveys) of the absolute value of the set size error, while Bias is the mean (signed) set size error.

**Performance depended on datasets and habitats.** We computed the F1-micro per dataset of the test set for eleven documented methods of the challenge (Figure 6). For all of them, we observed the same variations of performance across datasets: The performance was highest for the dataset "Inventaire Forestier IGN" (max=0.43 for *KDDI Ensemble*), it was much smaller for the dataset "National Plant Monitoring UK" (max=0.2 for *KDDI Ensemble*), but it was in general higher in the latter than in the two remaining datasets, namely "CBNMed" (max=0.16 for *KDDI Ensemble*) and "CBNA" (max=0.11 for *XG-Boost enviro. PA*). This is firstly explained by the different regional and habitat coverage of these datasets: "National Plant Monitoring UK" covers many habitats Great Britain, "Inventaire Forestier IGN" only covers french forest habitats, while "CBNMed" and "CBNA" cover respectively all habitats of the French mediterranean region and the French alpine region. As the "Inventaire Forestier IGN" is more habitat specific, the species composition is more homogeneous between the surveys compared to "CBNMed" and "CBNA", which are particularly diverse in plant habitats. Indeed, the "Inventaire Forestier IGN" shows a slower species accumulation when increasing the number of surveys (Figure 7), despite its much larger spatial extent. Besides, the difficulty of predicting the species composition is probably increased in the "National Plant Monitoring UK" given that the set size is compared to the "Inventaire Forestier IGN" (8.9 vs 19.8, Figure 7).

## 6. Discussion and Conclusion

For the first time in the GeoLifeCLEF series, we assembled a large test set of standardized presence-absence data, allowing to avoid the many evaluation biases due to the sampling issues of the more commonly available presence-only data, as noted the previous year [9]. We also provided a new type of remote sensing predictor, namely the satellite multi-band time-series.

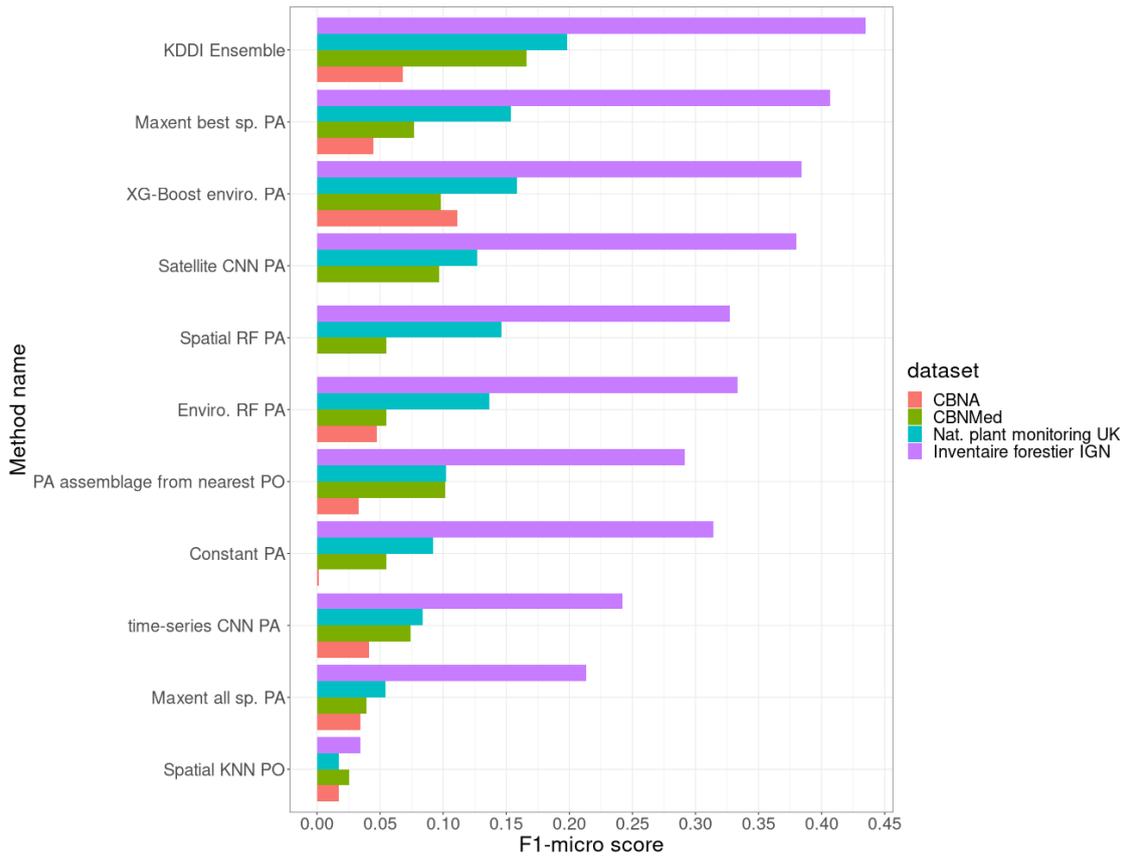

**Figure 6:** F1-micro score per method and PA dataset in GeoLifeCLEF 2023 test set.

The dataset assembled for the occasion forms a reference in the domain of species distribution modeling, in particular the problem of species composition prediction at very high spatial resolution and large extent, and will soon be published in a data paper ([12]).

GeoLifeCLEF has never offered so many possibilities, but has also never been so complex. This is probably one of the reasons why we observed a smaller participation than last year. Several participants acknowledged a likely significant room for improvement for their approaches given more time. Still, they developed very diverse methods using almost all types of predictors, and the KDDI research team even managed to combine the presence-only and presence-absence data to maximise their predictive performance to win the challenge [15], over-performing, by far, our best baseline.

We observed an evidently large room for improvement in our ability to exploit the rich information hidden in the masses of opportunistic PO data for predicting species composition in space. The challenge goes beyond this particular problem, and touches the proper integration of single positive labeled data for multi-label classification in machine learning ([18, 19]). We highlighted two central sources of errors arising from the bias in the way the single positive label is sampled, i.e. class detection biases and set size control. The true set size cannot be assessed from PO, because a record of a species does not inform us on the absence of the others, except if we can make further assumptions on the sampling process, as for instance by modeling the detection process (e.g. [20]). Furthermore, we cannot assume that the proportion of species observed in PO is constant for a given area, due to the tremendous variations in sampling effort across space [21]. Most participants methods significantly over-predicted set size, coherently with the optimization of main score (micro-F1) which favours larger predicted set for less accurate models. Nevertheless, it seems like many participants did not optimize the predicted set size. Further, there is likely a large room for improvement on the step of assemblage of discrete set predictions from quantitative model outputs to predict useful species compositions. We have also seen that the class detection bias is very strong in the PO data through a comparison with the objective PA data. This is likely due to the fact that in large opportunistic crowdsourcing programs (e.g. Pl@ntNet, iNaturalist), observers tend to focus on the most detectable, easy to identify and charismatic plant species. Set size variations and class detection bias can controlled by a calibration step on the PA data. The KDDI approach illustrated it (e.g. *KDDI PA/PO/PA* method) with their original learning scheme ([15]), optimizing a cross-entropy loss on the PO data, followed by an optimization of a binary-cross-entropy on the PA data, thus compensating for the class detection bias. Their predictions could be improved by accounting for set size variability. Indeed, one could capitalize on the set size information encoded in the species-wise presence probabilities, as for instance, the set size expectation.

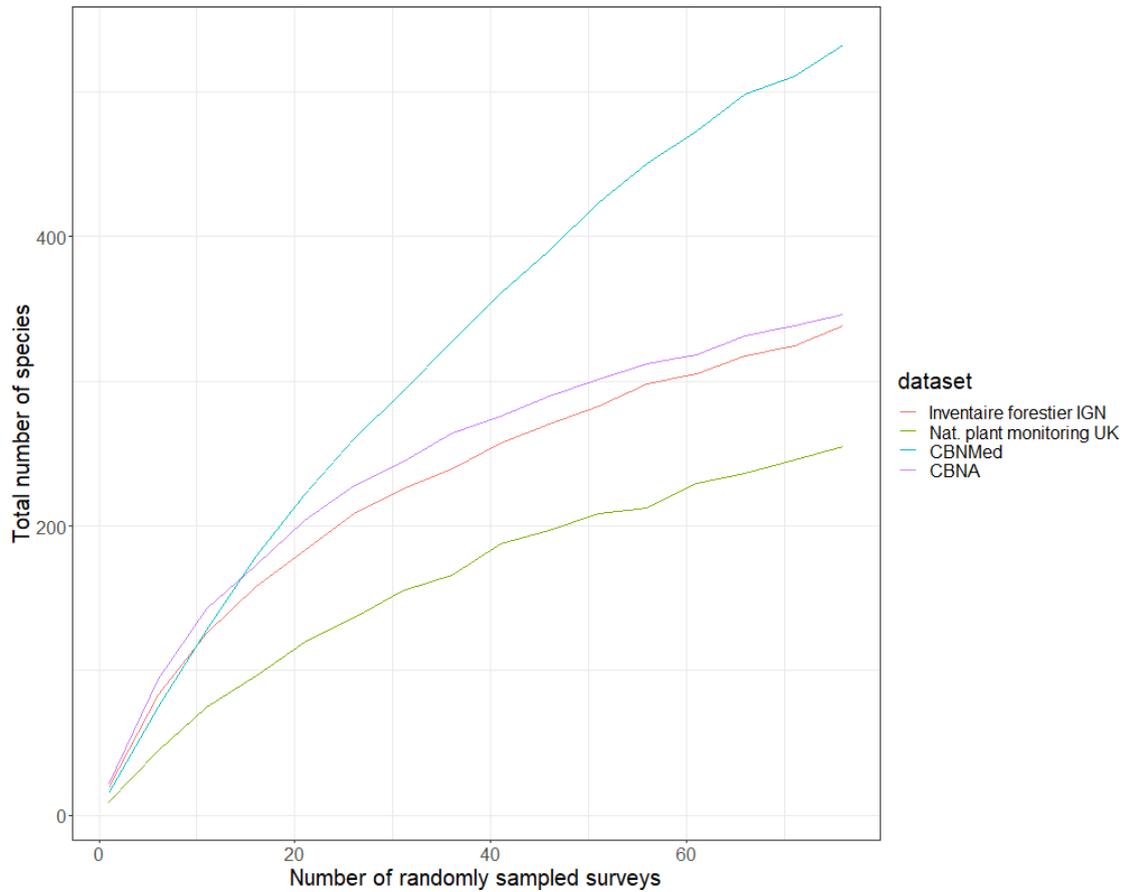

**Figure 7:** Number of species observed versus number of randomly selected surveys per dataset in the test set of GeoLifeCLEF2023.

Another important perspective is to better integrate the diverse available predictors whose formats and spatial or temporal resolutions are very different. Participants often built models based on one type of predictor, like satellite images (e.g. *Satellite CNN* of Quantmetry), time-series (e.g. *time-series CNN PA* of O. Youme) or tabular environmental variables (e.g. *XG-Boost enviro. PA* from Jiexun Xu). Yet, the KDDI team developed an efficient multi-modal CNN with late fusion of three CNNs respectively based on climatic rasters, soil rasters, and a stack of satellite images and the human footprint raster (*KDDI PA/PO/PA*), which improved compared to their climatic CNN alone. Besides, it would be interesting to evaluate the potential of this late fusion approach compared to a more simple ensembling method, or the potential further gain of end-to-end learning, which is generally a more difficult task. Anyways, not relying on a single type of predictor appears all the more important under predictor error, as the spatio-temporal patterns of error may differ from one predictor to the other. For instance, soil predictors are much less trustable for high latitudes [22], while cloudiness may obscure the information of satellite time-series in oceanic areas. Quantifying model predictive uncertainty and its variations across predictions may be a critical step towards a better integration of multi-modal models, as for instance through conformal prediction approaches.

Except the simple baseline *PA assemblage from nearest PO*, no method used species co-occurrences to our knowledge, namely the statistical relationship between species through space. This is surprising given the current emphasis about such approaches in the domain of species distribution modeling and ecology ([23]) and the existence of deep learning models using these properties ([24]).

In conclusion, an agenda for further methodological research could be decomposed into four main tasks: (i) integrating diverse predictors, particularly time-series of remote sensing or climatic variables, (ii) integrating diverse observation data types to correct sampling biases, for which the use of appropriate data model, through e.g. multi-head models, such as Poisson point processes have proven their worth ([10, 11]), (iii) addressing the aggregation of quantitative predictions into discrete set prediction, and (iv) better exploiting species relationships across space in deep learning models. We also note that, as the toolbox of deep species distribution model grows, it becomes more difficult to identify practices that consistently bring progress across contexts. Therefore, we once again emphasize the importance of experimental design, step-by-step complexification and the transparent description of results, in order to extract the best knowledge out of machine learning challenges.

## Acknowledgement

The research described in this paper was funded by the European Commission via the MAMBO (http:doi.org/10.3030/101060639) and GUARDEN (http:doi.org/10.3030/101060693) projects, which have received funding from the European Union's Horizon Europe research and innovation program under grant agreements 101060693 and 101060639.